\documentclass[10pt,a4paper]{IEEEtran}
\IEEEoverridecommandlockouts

\usepackage[utf8]{inputenc}
\usepackage[english]{babel}
\usepackage{cite}
\usepackage{graphicx}
\usepackage{amsmath,amsbsy,bm,amssymb,mathtools,dsfont}
\usepackage{algorithm,algpseudocode}
\usepackage{csquotes}
\usepackage{xspace}
\usepackage{pgfplots}
\usepackage{tabu,makecell,multirow}
\usepackage{caption}
\usepackage{subcaption}
\usepackage{placeins}
\usepackage[group-minimum-digits=3,separate-uncertainty]{siunitx}
\usepackage[hidelinks]{hyperref}
\usepackage{cleveref}

\crefname{figure}{Fig.}{Figs.}
\Crefname{figure}{Fig.}{Figs.}
\crefname{algorithm}{Alg.}{Algs.}
\Crefname{algorithm}{Algorithm}{Algorithms}
\crefname{table}{Tab.}{Tabs.}
\Crefname{table}{Tab.}{Tabs.}
\crefname{equation}{}{}
\Crefname{equation}{Equation}{Equations.}

\newcommand{\eg}{e.\,g.\xspace}
\newcommand{\cf}{cf.\xspace}
\newcommand{\quoting}[1]{``#1''}

\DeclareMathOperator\erf{erf}
\DeclareMathOperator*{\argmax}{arg\,max}
\DeclareMathOperator*{\argmin}{arg\,min}
\newcommand{\boldcal}[1]{\boldsymbol{\mathcal{#1}}}
\newcommand{\goesto}{\to}

\algnewcommand{\Not}{\textbf{not}\hspace{1ex}}

\newcommand{\timenum}[1]{\SI{#1}{\second}}

\pgfplotsset{compat=newest}
\usetikzlibrary{patterns}
\newenvironment{customlegend}[1][]{\begingroup\csname pgfplots@init@cleared@structures\endcsname\pgfplotsset{#1}}{\csname pgfplots@createlegend\endcsname \endgroup}\def\addlegendimage{\csname pgfplots@addlegendimage\endcsname}
\newcommand{\addlegendimageintext}[1]{\protect\tikz[]{\protect\begin{customlegend}[anchor=base,legend entries={\protect\empty},legend style={draw=none,inner sep=0pt,column sep=0pt,nodes={inner sep=0pt}}]\protect\addlegendimage{#1}\protect\end{customlegend}}}
\definecolor{colorgreen}{RGB}{77,175,74}
\definecolor{colorred}{RGB}{228,26,28}
\definecolor{colorblue}{RGB}{55,126,184}
\definecolor{colorgray}{RGB}{225,225,225}
\definecolor{colorlgreen}{RGB}{144,238,144}
\definecolor{colordgreen}{RGB}{0,238,144}

\begin{document}

\title{Adaptive Sampling of Pareto Frontiers with Binary Constraints Using Regression and Classification}
\author{%
\vspace{.3cm}%
\IEEEauthorblockN{Raoul Heese, Michael Bortz} \\
\IEEEauthorblockA{Fraunhofer Center for Machine Learning \\
Fraunhofer-Institut f{\"u}r Techno- und Wirtschaftsmathematik ITWM \\
Fraunhofer-Platz 1, 67663 Kaiserslautern, Germany\\
\{raoul.heese, michael.bortz\}@itwm.fraunhofer.de}%
\vspace{-.3cm}%
}%
\maketitle

\begin{abstract}
We present a novel adaptive optimization algorithm for black-box multi-objective optimization problems with binary constraints on the foundation of Bayes optimization. Our method is based on probabilistic regression and classification models, which act as a surrogate for the optimization goals and allow us to suggest multiple design points at once in each iteration. The proposed acquisition function is intuitively understandable and can be tuned to the demands of the problems at hand. We also present a novel ellipsoid truncation method to speed up the expected hypervolume calculation in a straightforward way for regression models with a normal probability density. We benchmark our approach with an evolutionary algorithm on multiple test problems.
\end{abstract}


\section{Introduction}
Bayesian optimization is a derivative-free strategy for the global optimization of computationally expensive black-box functions \cite{shahriari2016,frazier2018}. The basic idea is to build a surrogate for the objective function and to define an acquisition function based on this surrogate to decide where to sample next in an iterative approach. Ideally, a good compromise between the number of evaluations and the approximation quality of the solution can be found. Real-world optimization problems can also involve black-box constraints \cite{abdolshah2018} which restrict the solution space.\par
Multi-objective optimization (MOO) problems require the simultaneous optimization of more than one objective function \cite{miettinen1999,calendra2014}. Specifically, we consider MOO problems $\boldcal{O}$ of the form
\begin{align} \label{eqn:opt}
\underset{\mathbf{x}}{\text{minimize}} & \hspace{.5cm} \mathbf{y} \equiv \mathbf{y}(\mathbf{x}) \equiv (y_1(\mathbf{x}), \dots, y_n(\mathbf{x})) \nonumber \\
\text{subject to} & \hspace{.5cm} f \equiv f(\mathbf{x}) = \text{feasible} \nonumber \\
\text{where} & \hspace{.5cm} \mathbf{x} \in \boldcal{X} \subseteq \mathbb{R}^d \hspace{.15cm}\text{(design variables)} \nonumber \\
& \hspace{.5cm} \mathbf{y} \in \boldcal{Y} \subseteq \mathbb{R}^n \hspace{.15cm}\text{(objectives)} \nonumber \\
& \hspace{.5cm} f \in \mathcal{F} \equiv \{ \mathrm{feasible}, \mathrm{infeasible} \}
\end{align}
which also involve binary constraints $f(x)$.\par
For a non-trivial MOO problem there is no solution that allows an independent optimization of all objectives. Instead, trade-offs between two or more conflicting objectives have to be taken into account. In other words, there exist a set of Pareto optimal solutions for which none of the objective functions can be decreased without increasing another. We write $\mathbf{y} \preceq \mathbf{y'}$ to state that $\mathbf{y} \in \boldcal{Y}$ dominates $\mathbf{y'} \in \boldcal{Y}$, which means that
\begin{align} \label{eqn:pareto}
\mathbf{y} \preceq \mathbf{y'} \,\Leftrightarrow\, y_i \leq y_i' \,\forall\, i = 1,\dots,n \land \mathbf{y} \neq \mathbf{y'}.
\end{align}
The full set of Pareto optimal objectives
\begin{align} \label{eqn:pareto:set}
& \mathbf{P}(\boldcal{X},\mathbf{y}(\mathbf{x}),f(\mathbf{x})) \equiv \{ \mathbf{y} \in \boldcal{Y} \,|\, \,\exists\, \mathbf{x} \in \boldcal{X} \,:\, \mathbf{y} = \mathbf{y}(\mathbf{x}) \nonumber \\
& \hspace{.25cm} \land f(\mathbf{x}) = \mathrm{feasible} \land \mathbf{y'} \npreceq \mathbf{y} \,\forall\, \mathbf{x'} \in \boldcal{X} \setminus \{\mathbf{x}\} \,:\, \mathbf{y'} = \mathbf{y}(\mathbf{x'}) \nonumber \\
& \hspace{.25cm} \land f(\mathbf{x'}) = \mathrm{feasible} \,\}
\end{align}
consequently represents the solution of \cref{eqn:opt} and is also known as the Pareto frontier of the MOO problem.\par
Thus, our goal is to approximate this Pareto frontier as accurately as possible with as few evaluations
\begin{align} \label{eqn:S:eval}
\mathbf{S}(\mathbf{x}) \equiv (\mathbf{y}(\mathbf{x}),f(\mathbf{x})) = (\mathbf{y},f)
\end{align}
as possible. This function can, for example, represent a complex simulation with a very long runtime that does not converge at certain design points, which leads to an infeasible outcome. In particular, we assume to have no knowledge about the inner structure of $\mathbf{S}(\mathbf{x})$ and can only evaluate it in the sense of a black-box.\par
In this manuscript, we first present a novel method of adaptive optimization to solve such kind of problems. Subsequently, we benchmark our algorithm. Finally, we close with a short summary.
\section{Proposed method} \label{sec:method}
\begin{figure}[htb!]
\begin{center}
\includegraphics[scale=1]{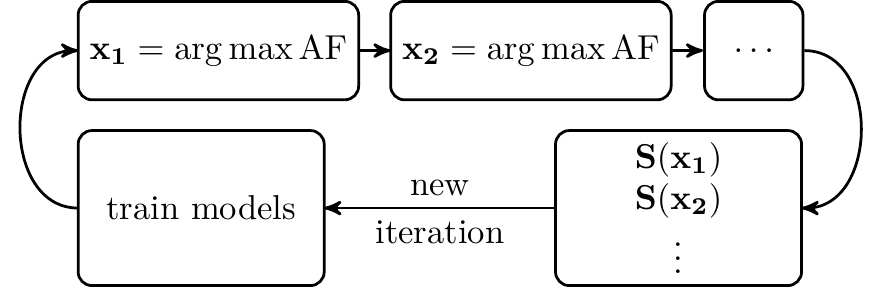}
\caption{Basic sketch of our adaptive optimization method which works along the lines of the well-known Bayesian optimization loop. First, the machine learning models, \cref{eqn:p}, are updated based on the currently available data. Second, a model-dependent acquisition function $\mathrm{AF}$, \cref{eqn:utility}, is maximized iteratively to obtain the design points $\mathbf{x_1}, \mathbf{x_2}, \dots$, which are then evaluated using \cref{eqn:S:eval}. For each maximization result, the acquisition function is changed based on the previous outcomes so that the following maximization will yield a different result. Finally, the loop may repeat again until a certain stopping criterion is reached. Our approach allows a parallelized evaluation of $N_{\mathrm{seq}}$ simulations, which leads to a significant reduction of the effective runtime.} \label{fig:utility-ploop}
\end{center}
\end{figure}

As sketched in \cref{fig:utility-ploop}, our proposed method of \emph{adaptive optimization} consists of three consecutive steps in the spirit of a typical Bayesian optimization loop, which repeats until a certain stopping criterion is reached. First, two machine learning models are trained which form a surrogate for $\mathbf{S}(\mathbf{x})$, \cref{eqn:S:eval}. Second, a model-dependent acquisition function $\mathrm{AF}(\mathbf{x})$ is maximized to obtain a design point of interest $\mathbf{x_1}$. This maximization may be repeated $N_{\mathrm{seq}}$ times to obtain a sequence of design points $\mathbf{x_1}, \mathbf{x_2}, \dots$, where the acquisition function is changed each time based on the previously obtained maximization results. And third, the sequence of design points is evaluated using $\mathbf{S}(\mathbf{x_1}), \mathbf{S}(\mathbf{x_2}), \dots$. A parallelized evaluation allows to reduce the total effective runtime of the algorithm in comparison with a single suggestion. In this sense, we approximate the Pareto frontier by adding new samples to our data set in each iteration. We have already demonstrated that a related approach can be used to explore the feasibility regions of a design space \cite{heese2019}. A Python implementation of our proposed method is provided in \cite{heese-adasamp-pareto}.

\subsection{Models}
Each evaluation of \cref{eqn:S:eval} yields a data point $\mathbf{d} \equiv ( \mathbf{x}, \mathbf{y}, f )$ and the set of $k$ of such data points is denoted as the data set
\begin{subequations} \label{eqn:D}
\begin{align}
\mathbf{D} \equiv \{ \mathbf{d_1}, \dots, \mathbf{d_k} \}
\end{align}
in the data space $\boldcal{D}$. Note that we use bold indices to iterate vector-valued elements of a set, and non-bold indices to indicate the respective vector components. Furthermore, we use subscripts and superscripts to denote the partial data sets
\begin{align}
\mathbf{D_x} & \equiv \{ \mathbf{x} \,\forall\, ( \mathbf{x}, \mathbf{y}, f ) \in \mathbf{D} \} \nonumber \\
	\mathbf{D_y^{\star}} & \equiv \{ \mathbf{y} \,\forall\, ( \mathbf{x}, \mathbf{y}, f ) \in \mathbf{D} \,|\, f = \text{feasible} \} \nonumber \\
	\mathbf{D_{xy}^{\star}} & \equiv \{ (\mathbf{x}, \mathbf{y}) \,\forall\, ( \mathbf{x}, \mathbf{y}, f ) \in \mathbf{D} \,|\, f = \text{feasible} \} \nonumber \\
	\mathbf{D_{xf}} & \equiv \{ ( \mathbf{x}, f) \,\forall\, ( \mathbf{x}, \mathbf{y}, f ) \in \mathbf{D} \}
\end{align}
\end{subequations}
with respect to $\mathbf{D}$.\par
To approximate \cref{eqn:S:eval} based on $\mathbf{D}$ we define two machine learning models. First, a probabilistic regression model
\begin{subequations} \label{eqn:p}
\begin{align} \label{eqn:py}
\hat{p}_y(\mathbf{y}|\mathbf{D_{xy}^{\star}},\mathbf{x}) : \boldcal{Y} \times \boldcal{D} \times \boldcal{X} \longmapsto \mathbb{R}_{\geq 0},
\end{align}
which predicts the probability density $\hat{p}_y$ of $\mathbf{S}(\mathbf{x})$ yielding the objectives $\mathbf{y}$. It is trained only on the feasible data points, whereas infeasible data points are ignored. Second, a probabilistic classification model
\begin{align} \label{eqn:pf}
\hat{p}_f(f | \mathbf{D_{xf}}, \mathbf{x}) : \boldcal{F} \times \boldcal{D} \times \boldcal{X} \longmapsto [0,1],
\end{align}
\end{subequations}
which predicts the probability $\hat{p}_f$ that evaluating $\mathbf{S}(\mathbf{x})$ leads to a feasibility $f$ and is trained on both feasible and infeasible data points.\par
These two models allow us to determine the expected optimization goal
\begin{subequations} \label{eqn:models}
\begin{align} \label{eqn:ey}
\mathbf{\hat{y}}(\mathbf{D_{xy}^{\star}}, \mathbf{x}) \equiv \int_{-\infty}^{\infty} \mathbf{y} \hat{p}_y(\mathbf{y}|\mathbf{D_{xy}^{\star}},\mathbf{x}) \mathbf{d y}
\end{align}
and the expected feasibility
\begin{align} \label{eqn:ef}
\hat{f}(\mathbf{D_{xf}}, \mathbf{x}) \equiv \sum_{f \in \boldcal{F}} f \hat{p}_f(f | \mathbf{D_{xf}}, \mathbf{x}),
\end{align}
\end{subequations}
respectively, which together represent a surrogate for \cref{eqn:S:eval}. For Bayes optimization one usually uses Gaussian process regression to model the objectives. For our method, however, any probabilistic regression model can be applied. For example, in \cref{sec:benchmark} we demonstrate the use of a Bayesian ridge regression model.

\subsection{Acquisition function}
Our model-dependent acquisition function
\begin{align} \label{eqn:utility}
\mathrm{AF}(\mathbf{x}) & \equiv \hphantom{+} U(\mathbf{w},\mathbf{D},\mathbf{y_{ref}},\gamma,\delta;\mathbf{x}) \nonumber \\
	& \equiv \hphantom{+} \frac{w_{\mathrm{opt}}}{|\mathbf{w}|_1} U_{\mathrm{opt}}(\mathbf{D},\mathbf{y_{ref}},\gamma;\mathbf{x}) + \frac{w_{\mathrm{con}}}{|\mathbf{w}|_1} U_{\mathrm{con}}(\mathbf{D};\mathbf{x}) \nonumber \\
	& \hphantom{\equiv} + \frac{w_{\mathrm{exp}}}{|\mathbf{w}|_1} U_{\mathrm{exp}}(\mathbf{D},\delta;\mathbf{x}) \in [0,1]
\end{align}
consists of three parts, the influence of which can be controlled by the choice of the weights $\mathbf{w} \equiv \left( w_{\mathrm{opt}}, w_{\mathrm{con}}, w_{\mathrm{exp}} \right) \geq 0$.\par
The optimization part $U_{\mathrm{opt}}$ ensures that the suggested design points improve the Pareto frontier within the feasible region, whereas the constraint-finding part $U_{\mathrm{con}}$ chooses points close to the border between feasible and infeasible domains to improve the classification model. Finally, the exploration part $U_{\mathrm{exp}}$ leads to a spreading of points in the design space and hence promotes an exploration of unknown regions. Summarized, \cref{eqn:utility} can be understood as an \emph{expected utility} of a design point $\mathbf{x}$ which takes all of the previously mentioned effects into account and tries to balance exploration and exploitation by the means of the chosen weights $\mathbf{w}$. In the following, we explain the three utility components in more detail.

\subsubsection{Optimization part}
The optimization utility
\begin{align} \label{eqn:utility:O}
U_{\mathrm{opt}}(\mathbf{D},\mathbf{y_{ref}},\gamma;\mathbf{x}) & \equiv \hat{p}_f(f = \text{feasible} | \mathbf{D_{xf}}, \mathbf{x}) \nonumber \\
& \hphantom{\equiv} \times O(\mathbf{D},\mathbf{y_{ref}},\gamma;\mathbf{x}),
\end{align}
consists of a product of two terms. The first term is based on the classification model, \cref{eqn:pf}, and corresponds to the predicted probability of a feasible outcome when evaluating $\mathbf{x}$. It acts as a weighting factor for the second term
\begin{align} \label{eqn:utility:O:O}
O(\mathbf{D},\mathbf{y_{ref},\gamma};\mathbf{x}) \equiv 1 - \exp \bigg[ \gamma \frac{\mathrm{EVI}(\mathbf{D_{xy}^{\star}},\mathbf{y_{ref}};\mathbf{x})}{-\Gamma(\mathbf{D_{y}^{\star}},\mathbf{y_{ref}})} \bigg],
\end{align}
which is a measure for the expected improvement of the Pareto frontier. It contains the user-defined parameter $\gamma \in \mathbb{R}_{>0}$ and the relative volume
\begin{align}
\Gamma(\mathbf{D_{y}^{\star}},\mathbf{y_{ref}}) \equiv \prod_{i=1}^{n} \max_{\mathbf{y} \in \mathbf{P}(\mathbf{D_y^{\star}})} \left( y_{\mathrm{ref},i} - y_i \right),
\end{align}
which both act as rescaling factors to the expected Pareto volume improvement
\begin{align} \label{eqn:utility:O:EVI}
\mathrm{EVI}(\mathbf{D_{xy}^{\star}},\mathbf{y_{ref}};\mathbf{x}) \equiv \int_{-\infty}^{+\infty} \Delta V(\mathbf{D_y^{\star}}, \mathbf{y_{ref}}; \mathbf{y}) \nonumber\\
\times \hat{p}_y(\mathbf{y}|\mathbf{D_{xy}^{\star}},\mathbf{x}) \mathbf{d y}
\end{align}
with
\begin{align}
\Delta V(\mathbf{D_y^{\star}}, \mathbf{y_{ref}}; \mathbf{y}) & \equiv V(\mathbf{D_y^{\star}} \cup \{ \mathbf{y} \}, \mathbf{y_{ref}}) \nonumber \\
& \hphantom{\equiv}\, - V(\mathbf{D_y^{\star}}, \mathbf{y_{ref}}),
\end{align}
where we have recalled the regression model, \cref{eqn:py}. Hence, \cref{eqn:utility:O:EVI} represents the expected increase of the Pareto volume
\begin{align} \label{eqn:paretovol}
V(\mathbf{D_y^{\star}}, \mathbf{y_{ref}}) \equiv \mathrm{Vol} ( \{ \mathbf{y} \in \mathbb{R}^n \,|\, \mathbf{P}(\mathbf{D_y^{\star}}) \preceq \mathbf{y} \preceq \mathbf{y_{ref}} \} )
\end{align}
with respect to a given reference point
\begin{align} \label{eqn:yref}
\mathbf{y_{ref}} \equiv \left( y_{\mathrm{ref},1}, \dots, y_{\mathrm{ref},n} \right) \in \mathbb{R}^n
\end{align}
when the point $\mathbf{y} \in \boldcal{Y}$ is added to the existing data set $\mathbf{D_y^{\star}}$. Here we make use of the Pareto optimal subset
\begin{align} \label{eqn:pareto:subset}
\mathbf{P}(\mathbf{D_y^{\star}}) \equiv \left\{ \mathbf{y} \in \mathbf{D_y^{\star}} \,|\, \mathbf{y'} \npreceq \mathbf{y} \,\forall\, \mathbf{y'} \in \mathbf{D_y^{\star}} \setminus \{\mathbf{y}\} \right\} \subseteq \mathbf{D_y^{\star}},
\end{align}
which is defined in analogy to \cref{eqn:pareto:set}. In \cref{sec:app:paretovol} we outline how \cref{eqn:utility:O:EVI} can be expressed in a closed form for a regression model with a normal probability density.\par
The expected Pareto volume improvement does not take the feasibility of the problem into account since the regression model assumes that all predicted points are feasible. However, the first factor in \cref{eqn:utility:O}, compensates this defect so that $U_{\mathrm{opt}}$ describes the expected improvement of the Pareto frontier weighted by the probability of the outcome being feasible. Similar expressions can also be found in \cite{gardner2014} in the context of Bayesian optimization with inequality constraints.

\subsubsection{Constraint-finding part}
The constraint-finding utility
\begin{align} \label{eqn:utility:S}
U_{\mathrm{con}}(\mathbf{D};\mathbf{x}) & \equiv p_{\nsucceq}(\mathbf{D_{xy}^{\star}};\mathbf{x}) S( \hat{p}_f(f = \text{feasible} | \mathbf{D_{xf}}, \mathbf{x}) )
\end{align}
also consists of a product of two terms. The first term
\begin{align} \label{eqn:utility:SR:pnd}
p_{\nsucceq}(\mathbf{D_{xy}^{\star}};\mathbf{x}) \equiv p(\hat{\mathbf{y}}(\mathbf{D_{xy}^{\star}};\mathbf{x}) \nsucceq \mathbf{y} \, \forall\, \mathbf{y} \in \mathbf{D_y^{\star}})
\end{align}
represents the probability that the expected optimization goal $\mathbf{\hat{y}}(\mathbf{D_{xy}^{\star}}, \mathbf{x})$, \cref{eqn:ey}, is not being dominated by the points in the already explored data set $\mathbf{D_{y}^{\star}}$. This probability can be expressed in a closed form for a regression model with a normal probability density as we outline in \cref{sec:app:nondomprob}. \Cref{eqn:utility:SR:pnd} acts as a weighting factor for the second term in \cref{eqn:utility:S}, which represents the Shannon information entropy of a binary event
\begin{align}
S(p) \equiv \frac{- p \ln p - (1-p) \ln (1-p) }{ \ln 2 }
\end{align}
for the predicted probability of a feasible outcome, \cref{eqn:pf}.\par
$S(p)$ attains its only maximum for $p=\frac{1}{2}$ so that the points with the highest predicted uncertainty are assigned the largest utility. The first term weights this value with the probability of the predicted objective being non-dominated so that design points which lead to expectably dominated predictions end up with a vanishing utility. $U_{\mathrm{con}}$ consequently describes the feasibility uncertainty weighted by the probability of the outcome not being dominated by already explored points.

\subsubsection{Explorative part}
The explorative utility
\begin{align} \label{eqn:utility:R}
U_{\mathrm{exp}}(\mathbf{D}, \delta; \mathbf{x}) \equiv p_{\nsucceq}(\mathbf{D_{xy}^{\star}};\mathbf{x}) R(\mathbf{D_x},\delta;\mathbf{x})
\end{align}
is again a product of two terms. The first term is the probability of the predicted objective being non-dominated, \cref{eqn:utility:SR:pnd}. The second term represents the normalized repulsion
\begin{align} \label{eqn:utility:R:R}
R(\mathbf{D_x},\delta;\mathbf{x}) \equiv \frac{\max_{\mathbf{x'} \in \mathbf{D_x}} \delta(\mathbf{x},\mathbf{x'})}{\max_{\mathbf{x'},\, \mathbf{x''} \in \boldcal{X}} \delta(\mathbf{x'},\mathbf{x''})}
\end{align}
based on a user-defined distance metric
\begin{align} \label{eqn:utility:R:d}
\delta : \boldcal{X} \times \boldcal{X} \longmapsto \mathbb{R}_{\geq 0}.
\end{align}
The denominator in \cref{eqn:utility:R:R} is a constant with respect to $\mathbf{x}$ and therefore just serves as a constant rescaling factor.\par
\Cref{eqn:utility:R:R} increases with an increasing distance of a design point $\mathbf{x}$ to already explored design points in $\mathbf{D_x}$. Hence, $U_{\mathrm{expl}}$ describes the point sparsity weighted by the probability of the outcome not being dominated by already explored points.

\subsection{Ellipsoid truncation method}
For our proposed method we have to repeatedly solve global optimization problems of the form
\begin{align} \label{eqn:max-utility}
\mathbf{x_k} \equiv \argmax_{\mathbf{x} \in \boldcal{X}} U(\mathbf{w},\mathbf{D},\mathbf{y_{ref}},\gamma,\delta;\mathbf{x})
\end{align}
to find the next point $\mathbf{x_k}$. Since global optimization strategies cannot guarantee an optimal solution, a suitable compromise between a high utility outcome and a low calculation time has to be found. In other words, since a numerical optimization result is an approximation, we can as well approximate the utility function to reduce computational effort. In \cref{sec:app:approxparetovol} we present the \emph{ellipsoid truncation method} as an approximation for the expected Pareto volume improvement $\mathrm{EVI} \approx \mathrm{\widetilde{EVI}}$, \cref{eqn:utility:O:EVI}, which can be applied for regression models with a normal probability density. The basic idea is to truncate all terms of little effect based on an intersection of the non-dominated regions with an ellipse centered at the predicted mean with an eccentricity proportional to the predicted standard deviation and the user-defined approximation control parameter $\sigma_{\mathrm{ref}} \in \mathbb{R}_{>0}$. Then
\begin{align} \label{eqn:utility:O:approx}
& U_{\mathrm{opt}}(\mathbf{D},\mathbf{y_{ref}},\gamma;\mathbf{x}) \approx \widetilde{U}_{\mathrm{opt}}(\mathbf{D},\mathbf{y_{ref}},\sigma_{\mathrm{ref}},\gamma;\mathbf{x}) \nonumber \\
	& \hspace{.75cm} \equiv \hat{p}_f(f = \text{feasible} | \mathbf{D_{xf}}, \mathbf{x}) \nonumber \\
	& \hspace{.75cm} \hphantom{\equiv} \times \Bigg\{ 1 - \exp \bigg[ \gamma \frac{\mathrm{\widetilde{EVI}}(\mathbf{D_{xy}^{\star}},\mathbf{y_{ref}}, \sigma_{\mathrm{ref}};\mathbf{x})}{-\Gamma(\mathbf{D_{y}^{\star}},\mathbf{y_{ref}})} \bigg] \Bigg\}
\end{align}
corresponds to an approximation $\mathbf{\widetilde{x_k}}(\sigma_{\mathrm{ref}}) \approx \mathbf{x_k}$ of the originally proposed expression, \cref{eqn:max-utility}, with $\mathbf{\widetilde{x_k}}(\sigma_{\mathrm{ref}} \goesto \infty) \goesto \mathbf{x_k}$.

\subsection{Algorithm} \label{sec:algorithm}
\Cref{alg:adaptiveoptimization} represents our proposed optimization method as sketched in \cref{fig:utility-ploop}. It contains the following functions:
\begin{itemize}
\item \textsc{Optimize}($\boldcal{O}, N_{\mathrm{seq}}$): Solve the black-box optimization task $\boldcal{O}$, \cref{eqn:opt}, using design point sequences of length $N_{\mathrm{seq}} \geq 1$. Return the resulting data set of pareto points, \cref{eqn:pareto:subset}, as an approximation of the true Pareto frontier, \cref{eqn:pareto:set}.
\item \textsc{InitialCalculation}($\boldcal{X}, \mathbf{S}$): Calculate an initial set of randomly chosen points on $\boldcal{X}$ using $\mathbf{S}$, \cref{eqn:S:eval}, and return the resulting data set $\mathbf{D}$, \cref{eqn:D},
\item \textsc{Stop}($\mathbf{D}$): Check whether a predefined stopping criterion is fulfilled (usually based on $\mathbf{D}$, \eg, a maximum number of sampled points) and return the boolean outcome.
\item \textsc{UpdateModels}($\mathbf{D}$): Train the machine learning models $\mathbf{M}$, \cref{eqn:p}, based on $\mathbf{D}$ and return them.
\item \textsc{Suggestion}($\boldcal{X}, \mathbf{M}, \mathbf{D'}$): Maximize the approximated utility function, \cref{eqn:utility,eqn:utility:O:approx}. The utility function is calculated using the machine learning models $\mathbf{M}$, \cref{eqn:p}, and the data set $\mathbf{D}'$. Return the resulting design point $\mathbf{x}$.
\item \textsc{Prediction}($\mathbf{M}, \mathbf{x}$): Evaluate the expectation values of the models $\mathbf{M}$ at the design point $\mathbf{x}$. Return both the expected objectives $\mathbf{\hat{y}}$, \cref{eqn:ey}, and the expected feasibility $\hat{f}$, \cref{eqn:ef}.
\item \textsc{Calculation}($\mathbf{S}, \mathbf{D'_x}$): Evaluate $\mathbf{S}$ for new design points $\mathbf{x} \in \mathbf{D'_x}$. Suggested points can be evaluated in parallel to reduce calculation time. Return the data set of results.
\item \textsc{Pareto}($\mathbf{D_y^{\star}}$): Return the Pareto optimal subset of $\mathbf{D_y^{\star}}$, \cref{eqn:pareto:subset}.	
\end{itemize}

\begin{algorithm}[htb!]
\caption{Adaptive optimization algorithm} \label{alg:adaptiveoptimization}
\begin{algorithmic}[1]
\Function{Optimize}{$\boldcal{O}, N_{\mathrm{seq}}$}
\State $\mathbf{D} \gets \Call{InitialCalculation}{\boldcal{X}, \mathbf{S}}$
\While{$\Not{} \Call{Stop}{\mathbf{D}}$}
\State $\mathbf{M} \gets \Call{UpdateModels}{\mathbf{D}}$
\State $\mathbf{D'} \gets \mathbf{D}$
\State $\nu \gets 0$
\While{$\nu < N_{\mathrm{seq}}$}
\State $\mathbf{x} \gets \Call{Suggestion}{\boldcal{X}, \mathbf{M}, \mathbf{D'}}$
\State $\mathbf{\hat{y}}, \hat{f} \gets \Call{Prediction}{\mathbf{M}, \mathbf{x}}$
\State $\mathbf{D'} \gets \mathbf{D'} \cup \{ ( \mathbf{x}, \mathbf{\hat{y}}, \hat{f} ) \}$
\State $\nu \gets \nu + 1$
\EndWhile
\State $\mathbf{D} \gets \mathbf{D} \cup \Call{Calculation}{\mathbf{S}, \mathbf{D'_x}}$
\EndWhile
\State \Return $\Call{Pareto}{\mathbf{D_y^{\star}}}$
\EndFunction
\end{algorithmic}
\end{algorithm}

\begin{table*}[htb!]
\caption{Definitions of the test problems used for our benchmarks with the notation from \cref{eqn:opt,eqn:f:c,sec:method}. Five of the six problems are also sketched in \cref{fig:bm:sketch}.}\label{tab:bm:problems}
{\tabulinesep=.75mm%
\resizebox{\linewidth}{!}{%
\begin{tabu}{|c||c|c|c|c|c|c|c|} \hline
\makecell{\textbf{Problem} \\ \textbf{name} \\ \textbf{and Ref.}} & \makecell{\textbf{Design} \\ \textbf{space} \\ $\boldcal{X}$} & \makecell{\textbf{Objectives} \\ $\mathbf{y}(\mathbf{x})$} & \makecell{\textbf{Constraints} \\ $\mathbf{c}(\mathbf{x})$} & \makecell{\textbf{Initial} \\ \textbf{domain}  \\ $\boldcal{X}_{\mathbf{0}}$} & \makecell{\textbf{Initial} \\ \textbf{samples} \\ $N_0$} & \makecell{\textbf{Reference} \\ \textbf{point} \\ $\mathbf{y_{ref}}$} & \makecell{\textbf{Adaptive} \\ \textbf{optimization}  \\ \textbf{parameters}} \\ \hline\hline
\makecell{BNH\\ \\ \cite{binh1997}} & \makecell{$d=2$ \\$x_1 \in [-5,15]$ \\ $x_2 \in [-10,10]$} & \makecell{$n=2$ \\ $y_1(\mathbf{x}) = 4 x_1^2 + 4 x_2^2$ \\ $y_2(\mathbf{x}) = (x_1 - 5)^2 + (x_2 - 5)^2$} & \makecell{$m=2$ \\ $c_1(\mathbf{x}) = (x_1 - 5)^2 + x_2^2 - 25$ \\ $c_2(\mathbf{x}) = - (x_1 - 8)^2 - (x_2 + 3)^2 + 7.7$} & \makecell{$x_1 \in [0,5]$ \\ $x_2 \in [-5,0]$} & $10$ & $(200,50)$ & \makecell{$\mathbf{w} = (0,1,0)$ \\ $\epsilon = 0$ \\ $\gamma = 10$ \\ $\sigma_{\mathrm{ref}} = 1$} \\ \hline
\makecell{SRN \\ \\ \cite{chankong2008}} & \makecell{$d=2$ \\ $x_1 \in [-20,20]$ \\ $x_2 \in [-20,20]$} & \makecell{$n=2$ \\ $y_1(\mathbf{x}) = 2 + (x_1-2)^2 + (x_2-1)^2$ \\ $y_2(\mathbf{x}) = 9 x_1 - (x_2-1)^2$} & \makecell{$m=2$ \\ $c_1(\mathbf{x}) = x_1^2 + x_2^2 - 255$ \\ $c_2(\mathbf{x}) = x_1 - 3 x_2 + 10$} & \makecell{$x_1 \in [0,20]$ \\ $x_2 \in [0,20]$} & $10$ & $(250,50)$ & \makecell{$\mathbf{w} = (0,1,0)$ \\ $\epsilon = 0$ \\ $\gamma = 10$ \\ $\sigma_{\mathrm{ref}} = 1$} \\ \hline
\makecell{OSY \\ \\ \cite{osyczka1995}} & \makecell{$d=6$ \\ $x_1 \in [0,10]$ \\ $x_2 \in [0,10]$ \\ $x_3 \in [1,5]$ \\ $x_4 \in [0,6]$ \\ $x_5 \in [1,5]$ \\ $x_6 \in [0,10]$} & \makecell{$n=2$ \\ $y_1(\mathbf{x}) = -25(x_1-2)^2 - (x_2-2)^2 \dots$ \\ \hfill $\dots - (x_3-1)^2  - (x_4-4)^2  \dots $ \\ \hfill $\dots - (x_5-1)^2$ \\ $y_2(\mathbf{x}) = x_1^2 + x_2^2 + x_3^2 + x_4^2 + x_5^2 + x_6^2$} & \makecell{$m=6$ \\ $c_1(\mathbf{x}) = -x_1 - x_2 +2$ \\ $c_2(\mathbf{x}) = x_1 + x_2 - 6$ \\ $c_3(\mathbf{x}) = x_2 - x_1 -2$ \\ $c_4(\mathbf{x}) = x_1 - 3 x_2 - 2$ \\ $c_5(\mathbf{x}) = (x_3-3)^2 + x_4 - 4$ \\ $c_6(\mathbf{x}) = - (x_5-3)^2 - x_6 + 4$} & \makecell{$x_1 \in [2,4]$ \\ $x_2 \in [0,3]$ \\ $x_3 \in [2,4]$ \\ $x_4 \in [0,2]$  \\ $x_5 \in [1,2]$ \\ $x_6 \in [0,10]$} & $100$ & $(0,80)$ & \makecell{$\mathbf{w} = (0,1,0)$ \\ $\epsilon = 0$ \\ $\gamma = 200$ \\ $\sigma_{\mathrm{ref}} = 5$} \\ \hline
\makecell{CEX \\ \\ \cite{fonseca1995} \\ (additional \\ constraints)} & \makecell{$d=2$ \\ $x_1 \in [0.1,1]$ \\ $x_2 \in [0,5]$} & \makecell{$n=2$ \\ $y_1(\mathbf{x}) = x_1$ \\ $y_2(\mathbf{x}) = \frac{x_2+1}{x_1}$} & \makecell{$m=4$ \\ $c_1(\mathbf{x}) = -9 x_1 - x_2 + 6$ \\ $c_2(\mathbf{x}) = - 9 x_1 + x_2 + 1$ \\ $c_3(\mathbf{x}) = .8 - x_1$ \\ $c_4(\mathbf{x}) = x_1 - \frac{2}{3}$ } & \makecell{$x_1 \in [.1,1]$ \\ $x_2 \in [2.5,.5]$} & $10$ & $(1,9)$ & \makecell{$\mathbf{w} = (1,3,1)$ \\ $\epsilon = 1$ \\ $\gamma = 1$ \\ $\sigma_{\mathrm{ref}} = 1.5$} \\ \hline
\makecell{FFF \\ \\ \cite{deb2001} \\ (additional \\ constraints)} & \makecell{$d=2$ \\ $x_1 \in [-1,1]$ \\ $x_2 \in [-1,1]$} & \makecell{$n=2$ \\ $y_1(\mathbf{x}) = 1 - \exp \left[ - (x_1 - \frac{1}{\sqrt{2}})^2 - (x_2 - \frac{1}{\sqrt{2}})^2 \right] $ \\ $y_2(\mathbf{x}) = 1 - \exp \left[ - (x_1 + \frac{1}{\sqrt{2}})^2 - (x_2 + \frac{1}{\sqrt{2}})^2 \right]$} & \makecell{$m=3$ \\ $c_1(\mathbf{x}) = x_1^2 + x_2^2 - \frac{1}{2}$ \\ $c_2(\mathbf{x}) = \min[ \{ y_1(\mathbf{x}) - .4, .6 - y_1(\mathbf{x}) \} ] $ \\ $c_3(\mathbf{x}) = \min[ \{ y_2(\mathbf{x}) - .4, .6 - y_2(\mathbf{x}) \} ]$} & \makecell{$x_1 \in [.25,1]$ \\ $x_2 \in [.25,1]$} & $10$ & $(1,1)$ & \makecell{$\mathbf{w} = (1,2,1)$ \\ $\epsilon = 1$ \\ $\gamma = 10$ \\ $\sigma_{\mathrm{ref}} = 1$} \\ \hline
\makecell{CIR \\ \\ (proposed \\ by us)} & \makecell{$d=2$ \\ $x_1 \in [-2,2]$ \\ $x_2 \in [-2,2]$} & \makecell{$n=2$ \\ $y_1(\mathbf{x}) = -\left( \frac{1}{2} \Theta[x_2-x_1] + x_1 \right)^2$ \\ $y_2(\mathbf{x}) = -\left( \frac{1}{2} \Theta[x_1-x_2] + x_2 \right)^2$} & \makecell{$m=1$ \\ $c_1(\mathbf{x}) = \min[ \{ (x_1-1)^2 + x_2^2 - 0.25 ,\! \dots$ \\ \hfill $\dots x_1^2 + (x_2-1)^2 - 0.25 \} ]$} & \makecell{$x_1 \in [.5,1.5]$ \\ $x_2 \in [-.5,.5]$} & $10$ & $(0,0)$ & \makecell{$\mathbf{w} = (1,1,1)$ \\ $\epsilon = 1$ \\ $\gamma = 1$ \\ $\sigma_{\mathrm{ref}} = 1$} \\ \hline
\end{tabu}%
}}%
\end{table*}

\begin{table*}[htb!]
\caption{Benchmark results for $\num{50}$ independent optimization runs with the notation from \cref{sec:performance metrics} for the test problems from \cref{tab:bm:problems}.}\label{tab:bm:results:nfe-tau}
{\tabulinesep=.75mm%
\resizebox{\linewidth}{!}{%
\begin{tabu}{|c||c|c|c|c||c|c|c|c|} \hline	
\multirow{2}{*}{\makecell{\textbf{Problem} \\ \textbf{name}}} & \multicolumn{4}{c||}{\textbf{Total number of evaluations} $N^{\delta}(\boldcal{O}, \texttt{adaptive-1}, \delta v)$} & \multicolumn{4}{c|}{\textbf{Break-even simulation times} $\tau(\boldcal{O}, \texttt{adaptive-1}, \texttt{nsgaii}, \delta v)$} \\ \cline{2-9}
& $\delta v = \num{.80}$ & $\delta v = \num{.85}$ & $\delta v = \num{.90}$ & $\delta v = \num{.95}$ & $\delta v = \num{.80}$ & $\delta v = \num{.85}$ & $\delta v = \num{.90}$ & $\delta v = \num{.95}$ \\ \hline\hline
BNH & \num{16.36 +- 2.54}       & \num{18.82 +- 2.56}       & \num{25.14 +- 4.89}       & \num{38.30 +- 5.40}       & \timenum{0.13 +- 0.08}         & \timenum{0.16 +- 0.08}         & \timenum{0.22 +- 0.10}         & \timenum{0.30 +- 0.09}         \\ \hline
SRN & \num{28.77 +- 23.26}      & \num{31.77 +- 23.08}      & \num{38.86 +- 22.54}      & \num{62.40 +- 15.71}      & \timenum{0.26 +- 0.20}         & \timenum{0.25 +- 0.16}         & \timenum{0.24 +- 0.12}         & \timenum{0.34 +- 0.13}         \\ \hline
OSY & \num{334.92 +- 121.41}    & \num{556.11 +- 172.00}    & \num{710.00 +- 112.00}    & $>\num{750}$              & \timenum{1.49 +- 1.34}         & \timenum{2.91 +- 2.37}         & \timenum{2.83 +- 1.56}         & --                             \\ \hline
CEX & \num{57.82 +- 41.50}      & \num{68.31 +- 42.04}      & \num{84.81 +- 37.48}      & \num{135.87 +- 37.88}     & \timenum{0.65 +- 0.76}         & \timenum{0.52 +- 0.52}         & \timenum{0.45 +- 0.33}         & \timenum{0.50 +- 0.28}         \\ \hline
FFF & \num{46.29 +- 24.81}      & \num{53.87 +- 27.50}      & \num{70.53 +- 26.49}      & \num{117.13 +- 20.78}     & \timenum{1.09 +- 1.17}         & \timenum{0.97 +- 1.19}         & \timenum{1.00 +- 1.02}         & \timenum{2.41 +- 1.05}         \\ \hline
CIR & \num{74.04 +- 13.49}      & \num{86.20 +- 14.22}      & \num{108.09 +- 17.41}     & \num{182.45 +- 15.37}     & \timenum{0.46 +- 0.23}         & \timenum{0.44 +- 0.21}         & \timenum{0.42 +- 0.22}         & \timenum{0.48 +- 0.16}         \\ \hline
\end{tabu}%
}}%
\end{table*}

\section{Benchmark} \label{sec:benchmark} 
\begin{figure}[htb!]
\begin{center}
\begin{subfigure}[b]{\linewidth}
	\centering\includegraphics[scale=1]{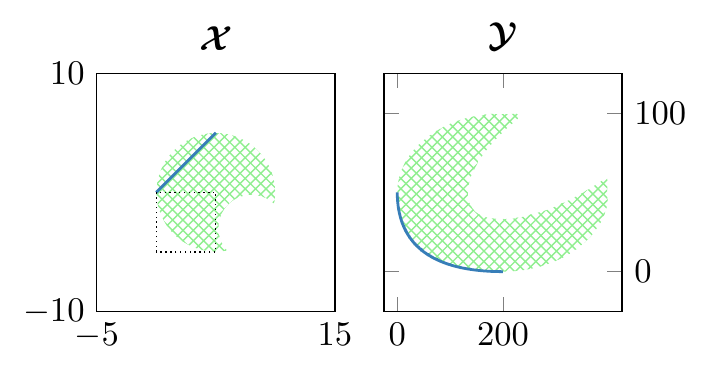}
	\caption{BNH}\label{fig:bm:sketch:bnh}
\end{subfigure}\\
\begin{subfigure}[b]{\linewidth}
	\centering\includegraphics[scale=1]{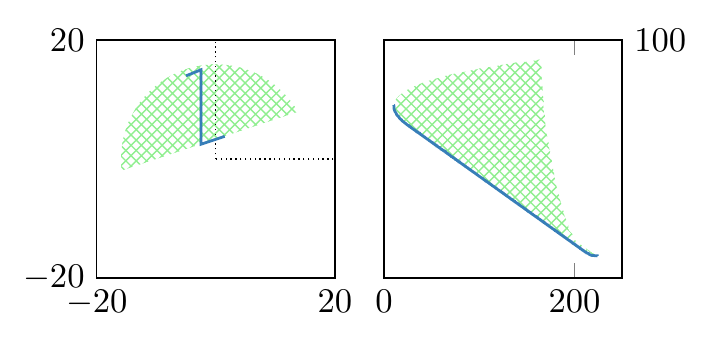}
	\caption{SRN}\label{fig:bm:sketch:srn}
\end{subfigure}\\
\begin{subfigure}[b]{\linewidth}
	\centering\includegraphics[scale=1]{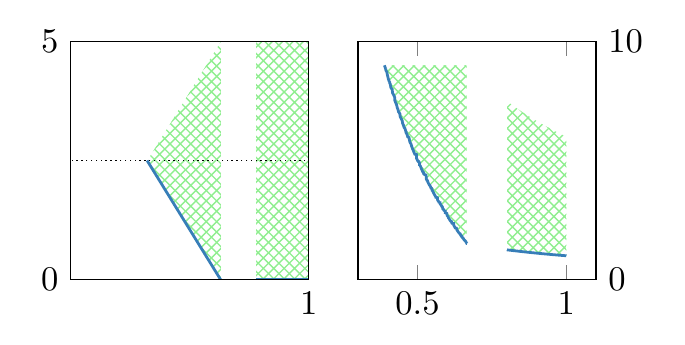}
	\caption{CEX}\label{fig:bm:sketch:cex}
\end{subfigure}\\
\begin{subfigure}[b]{\linewidth}
	\centering\includegraphics[scale=1]{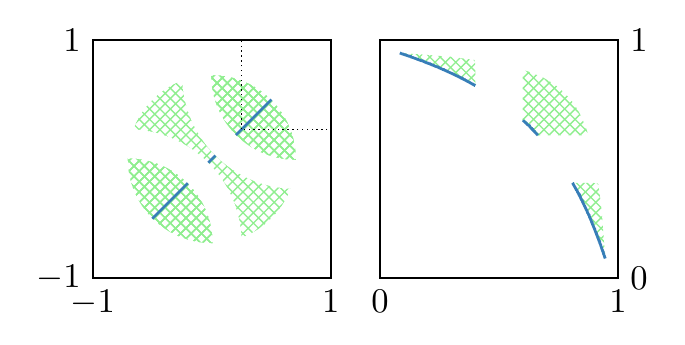}
	\caption{FFF}\label{fig:bm:sketch:fff}
\end{subfigure}\\
\begin{subfigure}[b]{\linewidth}
	\centering\includegraphics[scale=1]{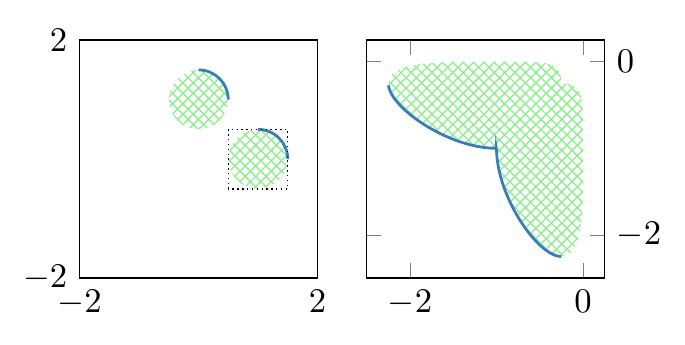}
	\caption{CIR}\label{fig:bm:sketch:cir}
\end{subfigure}
\end{center}
\caption{Sketches of five of the six test problems defined in \cref{tab:bm:problems}. For each problem, we show the design space $\boldcal{X}$ on the left and the objective space $\boldcal{Y}$ (together with its surrounding area) on the right with the following symbols: \protect\addlegendimageintext{area legend,pattern=crosshatch,pattern color=colorlgreen} true feasible region, \protect\raisebox{.075cm}{\addlegendimageintext{colorblue,thick,smooth}} true Pareto frontier, and \raisebox{.075cm}{\protect\addlegendimageintext{black,densely dotted}} border of $\boldcal{X}_{\mathbf{0}}$.}\label{fig:bm:sketch}
\end{figure}

In this section, we evaluate the performance of our proposed adaptive optimization algorithm $\texttt{adaptive}$ and compare it to the evolutionary algorithm $\texttt{nasgaii}$ \cite{deb2002}. Specifically, we study the six different benchmark problems listed in \cref{tab:bm:problems}, five of which are also sketched in \cref{fig:bm:sketch}. They are all in the form of \cref{eqn:opt}. Although we show constraints $\mathbf{c}(\mathbf{x})$ for each problem, we effectively use only the resulting binary feasibility 
\begin{align} \label{eqn:f:c}
f(\mathbf{x}) = \begin{cases} \mathrm{feasible} & \text{if} \, c_i(\mathbf{x}) \leq 0 \,\forall\, i= 1,\dots,m \\ \mathrm{infeasible} & \text{else} \end{cases}
\end{align}
as the mutual fulfillment condition of all $m$ constraints. In other words, we assume that the constraints are not accessible and we can only observe $f$. The Pareto frontiers of the problems CEX, FFF and CIR are disjointed in the design space and additionally disjointed in the objective space for the former two.\par
Initially, we randomly choose $N_0$ design points which are uniformly distributed on an initial design space $\boldcal{X}_{\mathbf{0}} \subseteq \boldcal{X}$. The resulting initial data set is then fed to the competing algorithms such that each candidate has the same initial information about the problem. For $\texttt{nasgaii}$ we choose a constant population size of $\num{50}$. The options for our proposed method are shown in the last column of \cref{tab:bm:problems}. We write $\texttt{adaptive-1}$ for the choice $N_{\mathrm{seq}} = 1$ and $\texttt{adaptive-5}$ for $N_{\mathrm{seq}} = 5$. As distance metric, \cref{eqn:utility:R:d}, we use
\begin{align} \label{eqn:bm:delta}
\delta_{\epsilon, \mathcal{X}}(\mathbf{x_1}, \mathbf{x_2}) \equiv 1 - \exp \left[ - \epsilon | \boldsymbol{\mu}_{\boldcal{X}}(\mathbf{x_1}) - \boldsymbol{\mu}_{\boldcal{X}}(\mathbf{x_2}) |_2^2 \right]
\end{align}
based on the user-defined control parameter $\epsilon \geq 0$ and the scaling function
\begin{subequations}
\begin{align}
\boldsymbol{\mu}_{\boldcal{X}}(\mathbf{x}) \equiv \frac{\mathbf{x} - \mathbf{x_{min}}(\boldcal{X})}{\mathbf{x_{max}}(\boldcal{X}) - \mathbf{x_{min}}(\boldcal{X})},
\end{align}
where
\begin{align}
\mathbf{x_{min}}(\boldcal{X}) \equiv \argmin_{\mathbf{x} \in \boldcal{X}} \mathbf{x}
\end{align}
and
\begin{align}
\mathbf{x_{max}}(\boldcal{X}) \equiv \argmax_{\mathbf{x} \in \boldcal{X}} \mathbf{x},
\end{align}
\end{subequations}
respectively. This choice of metric is motivated by the feature space distance of an exponential kernel \cite{schoelkopf2001}.\par
For the utility maximization, \cref{eqn:max-utility}, we use a two step approach. First, we perform a differential evolution \cite{storn1997} and then use an L-BFGS-B optimizer \cite{byrd1995} to further improve the result. The classification model, \cref{eqn:pf}, is realized with a RBF kernel support vector machine calibrated by Platt scaling and with hyperparameters optimized by cross-validation. For BNH, SRN, FFF, and CIR the regression model, \cref{eqn:py}, is a Gaussian process regression with Matern kernel, whereas for OSY and CEX we choose a Bayesian ridge regression model with polynomial features.

\subsection{Performance metrics} \label{sec:performance metrics}
We use three different metrics to quantify the performance of the competing algorithms, which we explain in the following.

\subsubsection{Relative total dominated volume}
The approximation quality of the Pareto frontier can be quantified by the relative total dominated volume
\begin{align} \label{eqn:bm:tdv}
\Delta V(\boldcal{O}, \mathbf{y_{ref}}, \texttt{alg}) \equiv \frac{V(\mathbf{D_y^{\star}}(\texttt{alg}), \mathbf{y_{ref}})}{V(\mathbf{P}(\boldcal{O}), \mathbf{y_{ref}})} \in [0,1]
\end{align}
as a ratio between the estimated Pareto volume and the volume of the actual Pareto set, where we have recalled \cref{eqn:paretovol}.

\subsubsection{Effective runtime}
We determine the pure runtime of the algorithm $T(\texttt{alg})$ from the start up to iteration $N_{\mathrm{iter}}(\texttt{alg})$ excluding the time for the simulation evaluations, \cref{eqn:S:eval}. Since for benchmark purposes we can freely choose an artificial evaluation time of the simulations to study different scenarios, we define that $N_{\mathrm{sim}}$ simulations can be evaluated in parallel during a constant runtime $T_{\mathrm{sim}}$. Given an algorithm that suggests sequences of $N_{\mathrm{seq}}(\texttt{alg})$ design points for each of $N_{\mathrm{iter}}(\texttt{alg})$ iterations, the total effective runtime of the algorithm is given by
\begin{align} \label{eqn:bm:T}
T_{\mathrm{eff}}(\texttt{alg}, N_{\mathrm{sim}}, T_{\mathrm{sim}}) & \equiv \hphantom{+} T_{\mathrm{sim}} \left\lceil \frac{ N_{\mathrm{seq}}(\texttt{alg}) }{ N_{\mathrm{sim}} } \right\rceil N_{\mathrm{iter}}(\texttt{alg}) \nonumber \\
& \hphantom{\equiv}+ T(\texttt{alg}).
\end{align}

\subsubsection{Break-even point}
The break-even simulation time between two competing algorithms $\texttt{adaptive-}N_{\mathrm{seq}}$ and $\texttt{nsgaii}$ represents the minimum value of $T_{\mathrm{sim}}$ for which $\texttt{adaptive-}N_{\mathrm{seq}}$ has a lower total effective runtime than $\texttt{nsgaii}$. For this purpose, we count the total number of iterations $N^{\delta}_{\mathrm{iter}}(\boldcal{O}, \texttt{alg}, \delta v)$ and the cumulated pure runtime $T^{\delta}(\boldcal{O}, \texttt{alg}, \delta v)$ (excluding the simulation time) of each algorithm to reach a certain Pareto approximation quality
\begin{align}
\Delta V(\boldcal{O}, \mathbf{y_{ref}}, \texttt{alg}) \geq \delta v
\end{align}
for a given relative total dominated volume $\delta v \in [0,1]$. This condition represents the stopping criterion \textsc{Stop}($\mathbf{D}$) for our method, \cf \cref{alg:adaptiveoptimization}.\par
If each algorithm suggests sequences of $N_{\mathrm{seq}}(\texttt{alg})$ design points for each of $N^{\delta}_{\mathrm{iter}}(\boldcal{O},\texttt{alg}, \delta v)$ iterations until this stopping criterion is reached and we set the number of parallelized simulation runs to $N_{\mathrm{sim}}=1$, then the break-even simulation time is given by
\begin{align} \label{eqn:bm:Tbe:single}
& \hphantom{\equiv} T_{\mathrm{sim}} > \tau(\boldcal{O}, \texttt{adaptive-}N_{\mathrm{seq}}, \texttt{nsgaii}, \delta v) \nonumber \\
& \equiv \frac{T^{\delta}(\boldcal{O},\! \texttt{adaptive-}N_{\mathrm{seq}},\! \delta v) \!-\! T^{\delta}(\boldcal{O},\! \texttt{nsgaii},\! \delta v)}{\nu^{\delta}(\texttt{nsgaii}) - \nu^{\delta}(\texttt{adaptive-}N_{\mathrm{seq}})}
\end{align}
with
\begin{align}
\nu^{\delta}(\texttt{alg}) \equiv N_{\mathrm{seq}}(\texttt{alg}) \times N^{\delta}_{\mathrm{iter}}(\boldcal{O},\texttt{alg}, \delta v).
\end{align}
Here we have assumed that $\texttt{adaptive-}N_{\mathrm{seq}}$ has a longer cumulated pure runtime, but a smaller value of $\nu^{\delta}(\texttt{alg})$ than $\texttt{nsgaii}$, which holds true in practice. 

\begin{figure}[htpb!]
\begin{center}
\includegraphics[scale=1]{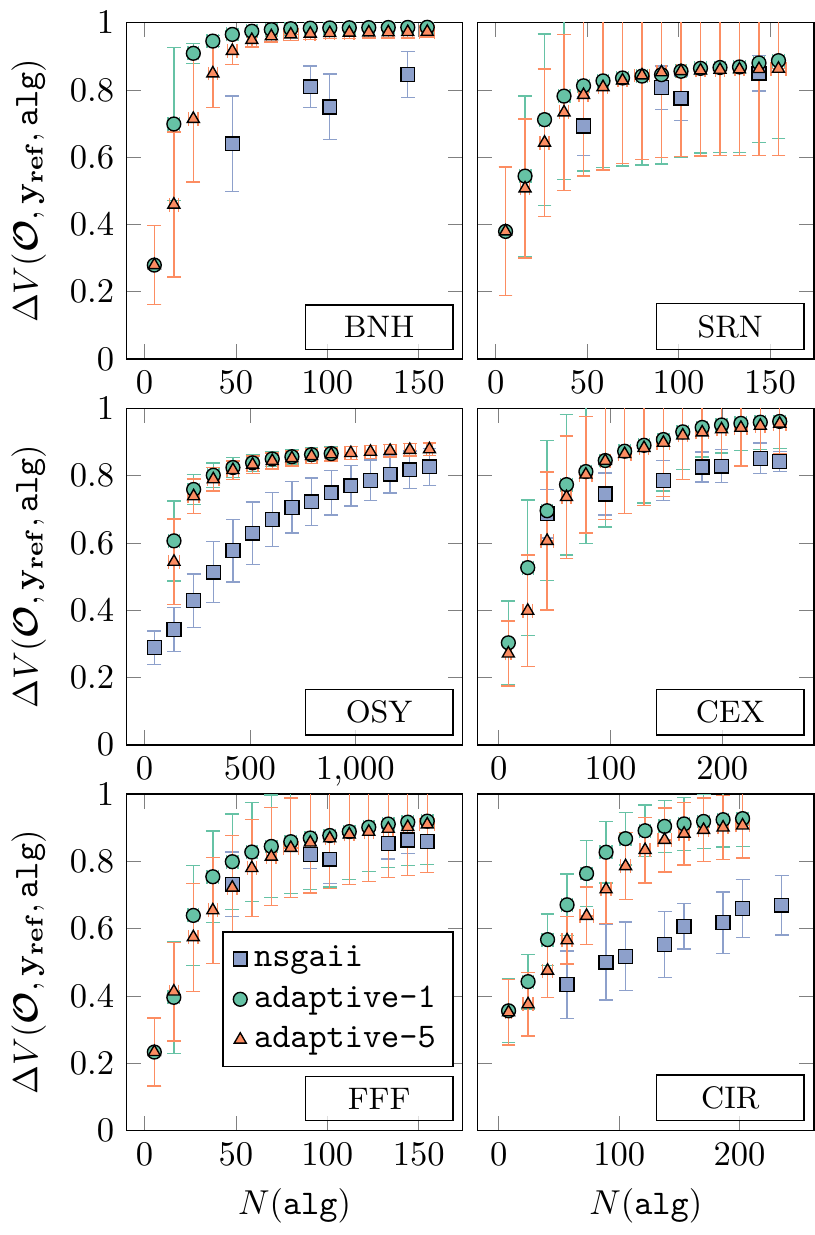}
\caption{Relative total dominated volume $\Delta V(\boldcal{O}, \mathbf{y_{ref}}, \texttt{alg})$, \cref{eqn:bm:tdv}, versus number of evaluated data points $N(\texttt{alg})$ of different algorithms $\texttt{alg}$. Each subplot corresponds to a different test problem $\boldcal{O}$ from \cref{tab:bm:problems} solved with $\num{50}$ independent runs. Steeper curves represent more effective optimization strategies.}\label{fig:results:nfe-tdv}
\end{center}
\end{figure}

\begin{figure}[htpb!]
\begin{center}
\includegraphics[scale=1]{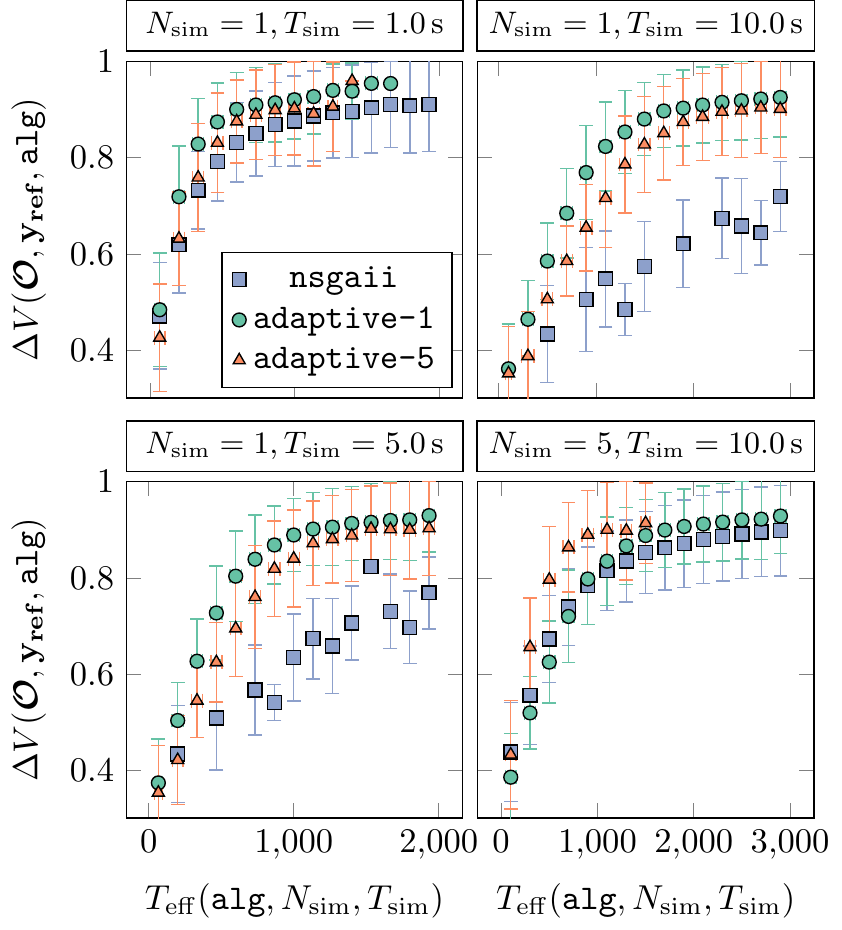}
\caption{Relative total dominated volume $\Delta V(\boldcal{O}, \mathbf{y_{ref}}, \texttt{alg})$, \cref{eqn:bm:tdv}, versus total effective runtimes $T_{\mathrm{eff}}(\texttt{alg}, N_{\mathrm{sim}}, T_{\mathrm{sim}})$, \cref{eqn:bm:T}, of different algorithms $\texttt{alg}$ for the test problem CIR from \cref{tab:bm:problems} solved with $\num{50}$ independent runs. In each subplot we assume a fixed number of simulation parallelizations $N_{\mathrm{sim}}$ and a fixed simulation evaluation time $T_{\mathrm{sim}}$. Steeper curves represent faster optimization strategies.} \label{fig:results:time-tdv:cir}
\end{center}
\end{figure}

\begin{figure}[htpb!]
\begin{center}
\begin{subfigure}[b]{\linewidth}
	\centering\includegraphics[scale=1]{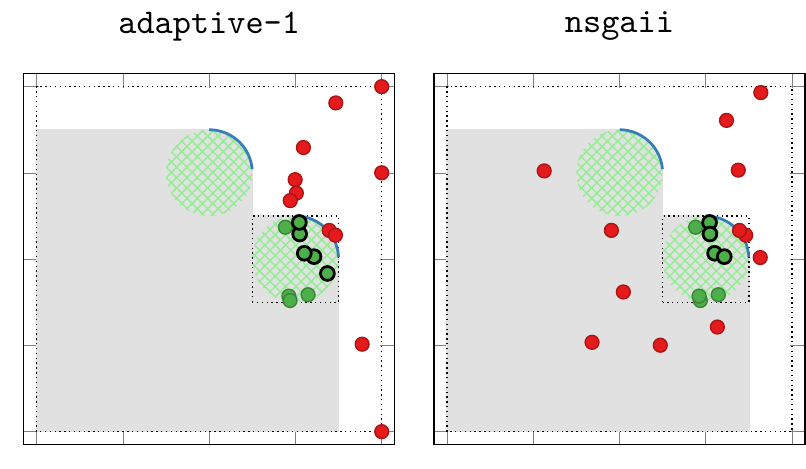}
	\caption{\num{20} suggested design points}\label{fig:results:sampling:cir:20}
\end{subfigure}\\\vspace{.35cm}
\begin{subfigure}[b]{\linewidth}
	\centering\includegraphics[scale=1]{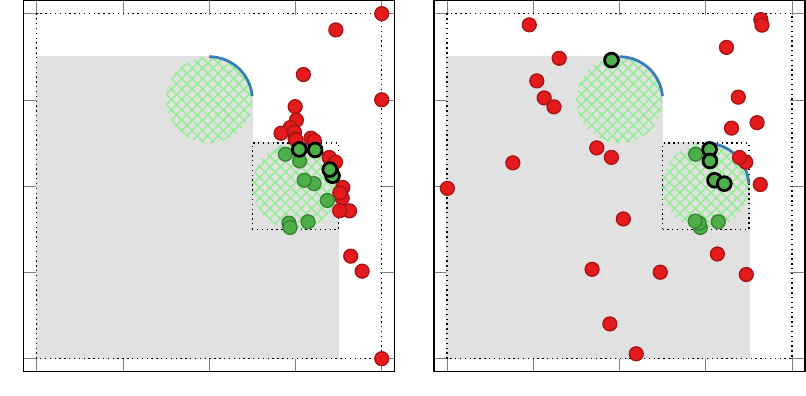}
	\caption{\num{35} suggested design points}\label{fig:results:sampling:cir:35}
\end{subfigure}\\\vspace{.35cm}
\begin{subfigure}[b]{\linewidth}
	\centering\includegraphics[scale=1]{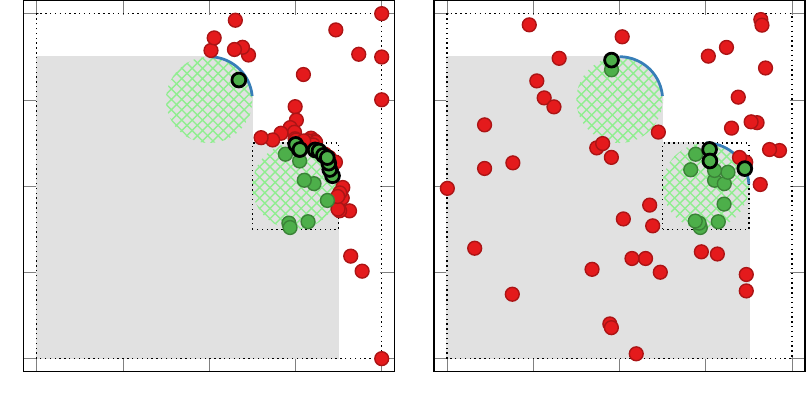}
	\caption{\num{60} suggested design points}\label{fig:results:sampling:cir:60}
\end{subfigure}\\\vspace{.35cm}
\begin{subfigure}[b]{\linewidth}
	\centering\includegraphics[scale=1]{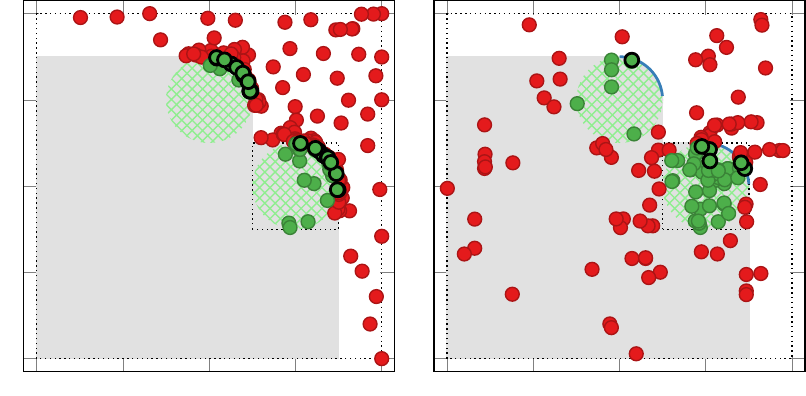}
	\caption{\num{160} suggested design points}\label{fig:results:sampling:cir:160}
\end{subfigure}
\end{center}
\caption{Exemplary visualization of different sampling stages for the test problem CIR from \cref{tab:bm:problems} based on the sketch from \cref{fig:bm:sketch:cir} with the following symbols: \protect\raisebox{.025cm}{\addlegendimageintext{draw=colorred!75!black,fill=colorred,only marks}} infeasible point, \protect\raisebox{.025cm}{\addlegendimageintext{draw=colorgreen!75!black,fill=colorgreen,only marks}} feasible point, \protect\raisebox{.025cm}{\addlegendimageintext{draw=black,thick,fill=colorgreen,only marks,mark=*}} Pareto optimal feasible point, \protect\raisebox{.075cm}{\addlegendimageintext{colorblue,thick,smooth}} true Pareto frontier, \protect\addlegendimageintext{area legend,fill=colorgray} true Pareto dominated region, \protect\addlegendimageintext{area legend,pattern=crosshatch,pattern color=colorlgreen} true feasible region, and \raisebox{.075cm}{\protect\addlegendimageintext{black,densely dotted}} borders of $\boldcal{X}_{\mathbf{0}}$ and $\boldcal{X}$. Both algorithms start with the same set of initial points. Clearly, $\texttt{adaptive-1}$ (left) chooses the data points more efficiently than $\texttt{nsgaii}$ (right) while still maintaining an explorative behavior.} \label{fig:results:sampling:cir}
\end{figure}

\subsection{Results}
For each test problem from \cref{tab:bm:problems} we run each algorithm $\num{50}$ times independently with different random seeds and evaluate the metrics discussed in \cref{sec:performance metrics}. For our proposed method we use the implementation from \cite{heese-adasamp-pareto} together with machine learning components from \cite{scikit-learn} and optimizers from \cite{scipy}, whereas for the evolutionary algorithm we use the implementation from \cite{platypus}.\par
In \cref{tab:bm:results:nfe-tau} we list the resulting total number of evaluations $N^{\delta}(\boldcal{O}, \texttt{adaptive-1}, \delta v)$ to reach the relative total dominated volume $\delta v$ and the break-even simulation times $\tau(\boldcal{O}, \texttt{adaptive-1}, \texttt{nsgaii}, \delta v)$ for which $\texttt{adaptive-1}$ runs faster than $\texttt{nsgaii}$. The table shows the mean values together with the respective standard deviations, where we make use of Gaussian error propagation for the deviations of $\tau$. For all test problems but the high-dimensional OSY (for which we stop prematurely after $\num{750}$ evaluations) we achieve a relative dominated volume of $\delta v=\num{0.95}$ with less than $\num{200}$ evaluations. Furthermore, the break-even simulation times are all of the order of seconds, even when we include the error interval of one standard deviation.\par
In \cref{fig:results:nfe-tdv} we show the number of evaluations $N(\texttt{alg})$ that are required to reach a certain relative dominated volume $\Delta V(\boldcal{O}, \mathbf{y_{ref}}, \texttt{alg})$. We plot mean values with error bars for the standard deviations, a steeper curve represents a more effective optimization strategy. We find that our algorithm is superior to the $\texttt{nsgaii}$ approach. It also becomes apparent that $\texttt{adaptive-1}$ is slightly better than $\texttt{adaptive-5}$, which means that a larger sequence of suggestions $N_{\mathrm{seq}}$ reduces the optimization quality. This is no surprise since further suggestions in the sequence beyond the first have to be made without additional information from evaluations.\par
However, longer sequence lengths allow to reduce the optimization time as we demonstrate in \cref{fig:results:time-tdv:cir}. Here we show the total effective runtimes $T_{\mathrm{eff}}(\texttt{alg}, N_{\mathrm{sim}}, T_{\mathrm{sim}})$ that are required to reach a certain relative dominated volume $\Delta V(\boldcal{O}, \mathbf{y_{ref}}, \texttt{alg})$ on the test problem CIR for different numbers of simulation parallelizations $N_{\mathrm{sim}}$ and different simulation evaluation times $T_{\mathrm{sim}}$. Clearly, for $N_{\mathrm{sim}}=5$, $\texttt{adaptive-5}$ is faster than $\texttt{adaptive-1}$ because its whole sequence of suggestions can be evaluated in parallel.\par
Finally, we explicitly show different sampling stages (\num{20}, \num{35}, \num{60}, and \num{160} suggested points) for a single optimization run on the test problem CIR in \cref{fig:results:sampling:cir}. We find that $\texttt{adaptive-1}$ achieves a much more efficient sampling which avoids Pareto dominated areas while maintaining an explorative behavior. This observation can already be made with a few samples, but becomes more and more obvious as the sampling progresses.
\section{Conclusion}
Summarized, we have presented a novel adaptive optimization algorithm on the foundation of Bayes optimization, which allows us to solve black-box multi-objective optimization problems with binary constraints. The weight-based utility function is intuitively understandable and can be tuned to the demands of the problems at hand. Our approach is based on probabilistic regression and classification models to predict the values and feasibility of the optimization objectives. Furthermore, we have made use of a novel ellipsoid truncation method to speed up our algorithm in a straightforward way. A benchmark has shown that our approach can compete with an evolutionary algorithm on a set of test problems with respect to the number of iterations and the calculation time.\par
In principle, our approach could also be used to optimize noisy simulations, which would require an appropriate modification of the machine learning models. Moreover, it could also be used to handle integer design variables, which would allow us to solve integer programming problems and mixed-integer programming problems. For this purpose, both the models and the optimization of the utility function had to be adapted accordingly. Another possible improvement would be the use of explicitly calculated gradients, which could greatly improve the performance of the optimization steps. All of these conceptional ideas could serve as a promising point of origin for further studies.

\section*{Acknowledgment}
We have realized our benchmark results with the help of \cite{scikit-learn,scipy,platypus}. This work was developed in the Fraunhofer Cluster of Excellence \quoting{Cognitive Internet Technologies}.

\bibliography{literature}{}
\appendix\subsection{Expected hypervolume improvement} \label{sec:app:paretovol}
In this section, we outline the calculation of \cref{eqn:utility:O:EVI}. Specifically, we use a straightforward approach along the lines of \cite{emmerich2008} with calculation time of order $O(|\mathbf{P}(\mathbf{D_y^{\star}})|^d)$. There are various approaches to speed up the calculation, see \eg, \cite{dachert2017,yang2019}.\par
\begin{figure}[htb!]
\begin{center}
\includegraphics[scale=1]{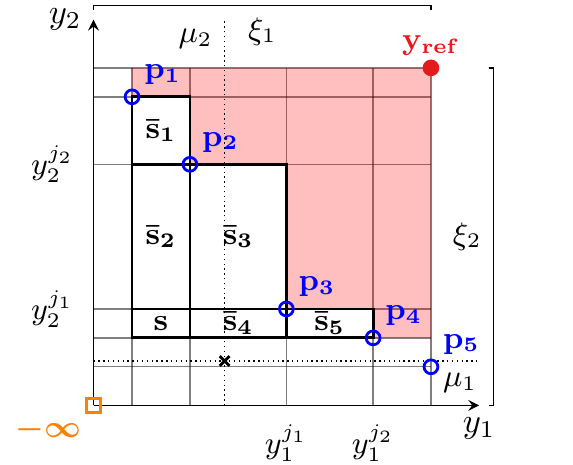}
\caption{Expected hypervolume improvement sketch ($n=2$) for \cref{sec:app:paretovol} with the following symbols: \protect\raisebox{.025cm}{\addlegendimageintext{thick,blue,only marks,mark=o}} Pareto optimal solutions, \protect\raisebox{.025cm}{\addlegendimageintext{thick,red,only marks,mark=*}} reference point $\mathbf{y_{ref}}$, \protect\raisebox{.025cm}{\addlegendimageintext{thick,orange,only marks,mark=square}} symbolic point in negative infinity $\boldsymbol{-\infty}$, \raisebox{.075cm}{\protect\addlegendimageintext{black,densely dotted}} means $\mu_1$ and $\mu_2$ of the regression model probability density intersecting at \protect\raisebox{.025cm}{\addlegendimageintext{thick,black,only marks,mark=x}}, and \protect\addlegendimageintext{area legend,fill=red!25!white} Pareto dominated region.}\label{fig:app:ehvi}
\end{center}
\end{figure}

As sketched in \cref{fig:app:ehvi}, we define the expected hypervolume improvement
\begin{align} \label{eqn:app:evi:def}
\mathrm{EVI}(\mathbf{D_{xy}^{\star}},\mathbf{y_{ref}};\mathbf{x}) & = \int_{-\infty}^{+\infty} \!\!\! \sum_{\mathbf{s} \in \mathbf{S_{\nsucceq}}(\mathbf{D_y^{\star}}, \mathbf{y_{ref}})} \hspace{-.5cm} \hat{p}_y(\mathbf{y}|\mathbf{D_{xy}^{\star}},\mathbf{x}) \nonumber \\
	& \hphantom{=} \times \Delta V({\mathbf{s}}, \mathbf{S_{\nsucceq}}(\mathbf{D_y^{\star}}, \mathbf{y_{ref}}); \mathbf{y})  \mathbf{d y}
\end{align}
locally on a non-regular grid of sectors
\begin{align}
\mathbf{S}(\mathbf{D_y^{\star}}, \mathbf{y_{ref}}) & \equiv \Big\{ \mathbf{s} = ((y_1^{j_1},y_1^{j_2}),\dots,(y_n^{j_n},y_n^{j_{n+1}})) \in \boldsymbol{\Xi} \,|\, \nonumber \\
& \hphantom{\equiv\Big\{}\,\, y_i^{j_i} = \begin{rcases} \begin{dcases} {y_{\mathrm{ref}}}_i & \text{if}\, j_i = 0 \\ -\infty & \text{if}\, j_i > | \mathbf{P}(\mathbf{D_y^{\star}}) | \\ y_i \in \mathbf{P}(\mathbf{D_y^{\star}}) & \text{else} \end{dcases} \end{rcases} \nonumber \\
& \hphantom{\equiv\Big\{}\,\, \forall\, j_i \in \{ 0, \dots, | \mathbf{P}(\mathbf{D_y^{\star}}) |+1 \} \nonumber \\
& \hphantom{\equiv\Big\{}\,\, \forall\, i = 1, \dots, n \nonumber \\
& \hphantom{\equiv\Big\{}\,\, \land y_i^{j_{i-1}} \geq y_i^{j_i} \geq y_i^{j_{i+1}} \nonumber \\
& \hphantom{\equiv\Big\{}\,\, \forall\, j_i \in \{ 1, \dots, | \mathbf{P}(\mathbf{D_y^{\star}}) | \} \nonumber \\
& \hphantom{\equiv\Big\{}\,\, \forall\, i = 1, \dots, n  \Big\}.
\end{align}
Each grid sector is defined as a tuple of corner points $(y_1^{j_1},y_1^{j_2}),\dots,(y_n^{j_n},y_n^{j_{n+1}})$ from the set of possible grid points
\begin{align}
\boldsymbol{\Xi} \equiv \xi_1^2 \times \cdots \times \xi_n^2,
\end{align}
where
\begin{align}
\xi_i \equiv & \{ y_i \in \mathbb{R} \cup \{ -\infty \} \,|\, \mathbf{y} \in \mathbf{P}(\mathbf{D_y^{\star}}) \cup \{ \mathbf{y_{ref}}, \boldsymbol{-\infty} \} \} \nonumber \\
	& \,\forall\, i = 1 \dots n.
\end{align}
For convenience we use $\boldsymbol{-\infty}$ to denote a vector whose entries are all negative infinity. The grid expands from this symbolic point to the reference point $\mathbf{y_{ref}}$, \cref{eqn:yref}.\par
Furthermore, we have made use of the subset
\begin{align}
\mathbf{S_{\nsucceq}}(\mathbf{D_y^{\star}}, \mathbf{y_{ref}}) & \equiv \Big\{ \mathbf{s} = ((y_1^{j_1},y_1^{j_2}),\dots,(y_n^{j_n},y_n^{j_{n+1}})) \in \boldsymbol{\Xi} \,|\, \nonumber \\
& \hphantom{\equiv\Big\{}\,\, \mathbf{s} \in \mathbf{S}(\mathbf{D_y^{\star}}, \mathbf{y_{ref}}) \nonumber \\
& \hphantom{\equiv\Big\{}\,\, \land \,\forall\, \mathbf{y} \in \mathbf{P}(\mathbf{D_y^{\star}}) \,\exists\, i \,:\, y_i^{j_i} < y_i \Big\}
\end{align}
of non-dominated grid sectors of $\mathbf{S}(\mathbf{D_y^{\star}}, \mathbf{y_{ref}})$ and the local volume integrand
\begin{align}
\Delta V(\mathbf{s}, \mathbf{S}; \mathbf{y}) & \equiv \!\!\!\!\!\! \sum_{\mathbf{\bar{s}} \in \mathbf{S_L}(\mathbf{s}, \mathbf{S})} \,\, \prod_{i=1}^n (y_i - \delta_{\bar{y}_i^{j_i}, y_i^{j_i}} \bar{y}_i^{j_{i+1}}) \mathds{1}_V(\mathbf{s}; \mathbf{y})
\end{align}
with the indicator function
\begin{align}
\mathds{1}_V(\mathbf{s}; \mathbf{y}) \equiv \begin{cases} 1 & \text{if}\, y_i^{j_{i+1}} \leq y_i \leq y_i^{j_i} \,\forall\,i \\ 0 & \text{otherwise}\end{cases}
\end{align}
as well as the set of local sectors
\begin{align}
\mathbf{S_L}(\mathbf{s}, \mathbf{S}) & \equiv \Big\{ \mathbf{\bar{s}} = ((\bar{y}_1^{j_1},\bar{y}_1^{j_2}),\dots,(\bar{y}_n^{j_n},\bar{y}_n^{j_{n+1}})) \in \boldsymbol{\Xi} \,|\, \mathbf{\bar{s}} \in \mathbf{S} \nonumber \\
& \hphantom{\equiv\Big\{}\,\ \land \exists i : \bar{y}_i^{j_i} \!\geq\! y_i^{j_i} \nonumber \\
& \hphantom{\equiv\Big\{}\,\ \text{with}\,\,\, \mathbf{s} \!=\! ((y_1^{j_1},y_1^{j_2}), \dots, (y_n^{j_n},y_n^{j_{n+1}})) \Big\},
\end{align}
which is a subset of $\mathbf{S}$.\par
For a regression model, \cref{eqn:py}, with separable probabilites 
\begin{align} \label{eqn:app:py:separable}
\hat{p}_y(\mathbf{y}|\mathbf{D_{xy}^{\star}},\mathbf{x}) = \prod_{i=1}^n \hat{p}_{y_i}(y_i|\mathbf{D_{xy}^{\star}},\mathbf{x})
\end{align}
we can interchange the integral and sums in \cref{eqn:app:evi:def} due to Tonelli's theorem and arrive at
\begin{align} \label{eqn:app:evi}
& \mathrm{EVI}(\mathbf{D_{xy}^{\star}},\mathbf{y_{ref}};\mathbf{x}) = \!\!\! \sum_{\mathbf{s} \in \mathbf{S_{\nsucceq}}(\mathbf{D_y^{\star}}, \mathbf{y_{ref}})} \,\,\, \sum_{\mathbf{\bar{s}} \in \mathbf{S_L}(\mathbf{s}, \mathbf{S_{\nsucceq}}(\mathbf{D_y^{\star}}, \mathbf{y_{ref}}))} \nonumber \\
& \times \prod_{i=1}^n \int_{\bar{y}_i^{j_{i+1}}}^{\bar{y}_i^{j_i}} (y_i - \delta_{\bar{y}_i^{j_i}, y_i^{j_i}} \bar{y}_i^{j_{i+1}}) \hat{p}_{y_i}(y_i|\mathbf{D_{xy}^{\star}},\mathbf{x}) \mathrm{d} y_i.
\end{align}
In case of normal probability densities
\begin{align} \label{eqn:pv:Gauss}
\hat{p}_{y_i}(y_i | \mathbf{D_{xy}^{\star}},\mathbf{x}) = \mathcal{N}(y_i | \mu_i(\mathbf{D_{xy}^{\star}},\mathbf{x}), \sigma_i^2(\mathbf{D_{xy}^{\star}},\mathbf{x}))
\end{align}
with mean $\mu_i(\mathbf{D_{xy}^{\star}},\mathbf{x}) \equiv \mu_i$ and standard deviation $\sigma_i(\mathbf{D_{xy}^{\star}},\mathbf{x}) \equiv \sigma_i$ for $i=1,\dots,n$, this expression can be straightforwardly written in a closed form using exponential and error functions. The closed form contains only three different types of integrals $I_1$, $I_2$, and $I_3$ with
\begin{subequations} \label{eqn:I}
\begin{align}
& I_1(a,b,c;\mu,\sigma) \equiv \int_{a}^{b} (y - c) \mathcal{N}(y | \mu, \sigma^2) \mathrm{d} y \nonumber \\
& = \frac{\mu - c}{2} \left[ \erf \left( \frac{b - \mu}{\sqrt{2} \sigma} \right) - \erf \left( \frac{a - \mu}{\sqrt{2} \sigma} \right)\right] \nonumber \\
& \hphantom{=} +\! \frac{\sigma}{\sqrt{2 \pi}} \left[ \exp \left( \!- \frac{(a - \mu)^2}{2 \sigma^2} \right) - \exp \left( \!- \frac{(b - \mu)^2}{2 \sigma^2} \right) \right],
\end{align}
\begin{align}
I_2(b,c;\mu,\sigma) & \equiv \int_{-\infty}^{b} (y - c) \mathcal{N}(y | \mu, \sigma^2) \mathrm{d} y \nonumber \\
& = \frac{\mu - c}{2} \left[ \erf \left( \frac{b - \mu}{\sqrt{2} \sigma} \right) + 1 \right] \nonumber \\
& \hphantom{=} + \frac{\sigma}{\sqrt{2 \pi}} \exp \left( - \frac{(b - \mu)^2}{2 \sigma^2} \right),
\end{align}
and
\begin{align}
I_3(a,b;\mu,\sigma) & \equiv \int_{a}^{b} y \mathcal{N}(y | \mu, \sigma^2) \mathrm{d} y \nonumber \\
& = \frac{1}{2} \left[ \erf \left( \frac{b - \mu}{\sqrt{2} \sigma} \right) - \erf \left( \frac{a - \mu}{\sqrt{2} \sigma} \right)\right],
\end{align}
\end{subequations}
respectively. Here we have used the abbreviations $\mu \in \left\{ \mu_1, \dots, \mu_n \right\}$, $\sigma \in \left\{ \sigma_1, \dots, \sigma_n \right\}$, and $a, b, c \in \xi_i \setminus \{ \boldsymbol{-\infty} \} \,\forall\, i = 1, \dots, n$.\par
In practice, the standard deviations $\sigma_i$ can become smaller than the numerical precision. In this case, the limit
\begin{align} \label{eqn:pv:Gauss:sigma:vanish}
& \mathcal{N}(y_i | \mu_i, \sigma_i^2) \xrightarrow{\sigma_i \goesto 0} \delta(y_i - \mu_i)
\end{align}
with the Dirac delta distribution $\delta$ can be use to obtain
\begin{subequations} \label{eqn:I:delta}
\begin{align}
I_1(a,b,c;\mu,\sigma) \xrightarrow{\sigma \goesto 0} (\mu - c) \Theta \left[ b - \mu \right]  \Theta \left[ \mu - a \right],
\end{align}
\begin{align}
I_2(b,c;\mu,\sigma) \xrightarrow{\sigma \goesto 0} (\mu - c) \Theta \left[ b - \mu \right],
\end{align}
and
\begin{align}
I_3(a,b;\mu,\sigma) \xrightarrow{\sigma \goesto 0} \Theta \left[ b - \mu \right] \Theta \left[ \mu - a \right],
\end{align}
\end{subequations}
respectively, where $\Theta$ denotes the Heaviside theta function.

\subsection{Ellipsoid truncation method} \label{sec:app:approxparetovol}
\begin{figure}[htb!]
\begin{center}
\includegraphics[scale=1]{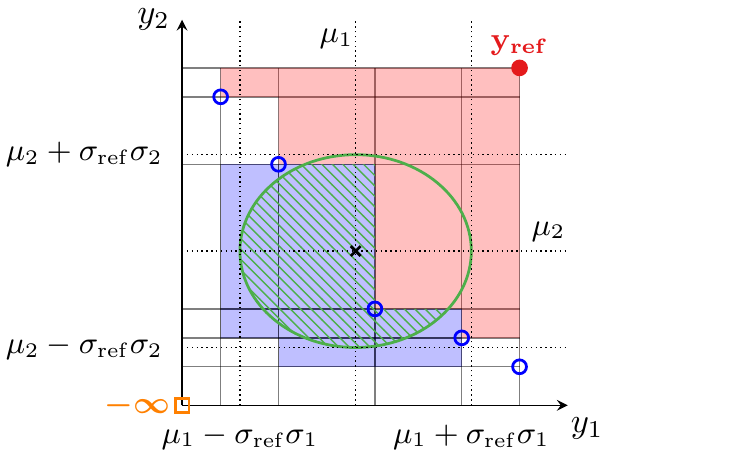}
\caption{Ellipsoid truncation method sketch ($n=2$) for \cref{sec:app:approxparetovol} with the symbols from \cref{fig:app:ehvi} and the additional symbols: \protect\addlegendimageintext{area legend,fill=blue!25!white} non-truncated areas, which partially \protect\addlegendimageintext{area legend,pattern=north west lines,pattern color=colorgreen} overlap with the \protect\raisebox{.075cm}{\addlegendimageintext{thick,colorgreen}} truncation ellipse.}\label{fig:app:ellipse}
\end{center}
\end{figure}

In this section, we outline the expected hypervolume approximation used in \cref{eqn:utility:O:approx}, which is applicable for a regression model with a normal and separable probability density, \cref{eqn:py,eqn:app:py:separable,eqn:pv:Gauss}, with means $\mu_1,\dots,\mu_n$ and standard deviations $\sigma_1,\dots,\sigma_n$. As sketched in \cref{fig:app:ellipse}, the basic idea is to neglect sectors in \cref{eqn:app:evi} which do not intersect with an ellipse of a given size corresponding to the probability density function so that
\begin{align}
\mathrm{EVI}(\mathbf{D_{xy}^{\star}},\mathbf{y_{ref}};\mathbf{x}) \!\approx\! \mathrm{\widetilde{EVI}}(\mathbf{D_{xy}^{\star}},\mathbf{y_{ref}},\sigma_{\mathrm{ref}}; \mathbf{x}) \!\equiv\! \hspace{-1.2cm} \sum_{\mathbf{s} \in \mathbf{\widetilde{S}_{\nsucceq}}(\mathbf{D_y^{\star}}, \mathbf{y_{ref}}, \sigma_{\mathrm{ref}} ;\mathbf{x})} \hspace{-1cm} \dots
\end{align}
with the subset
\begin{align}
\mathbf{\widetilde{S}_{\nsucceq}}(\mathbf{D_y^{\star}}, \mathbf{y_{ref}}, \sigma_{\mathrm{ref}}; \mathbf{x}) & \equiv \Big\{ \mathbf{s} = ((y_1^{j_1},y_1^{j_2}),\dots,(y_n^{j_n},y_n^{j_{n+1}})) \nonumber \\
& \hphantom{\equiv\Big\{}\,\ \!\in\! \boldsymbol{\Xi} \,|\, \mathbf{s} \!\in\! \mathbf{S_{\nsucceq}}(\mathbf{D_y^{\star}}, \mathbf{y_{ref}}) \nonumber \\
& \hphantom{\equiv\Big\{}\,\, \hspace{-.6cm} \land \mathbf{H}(\mathbf{s}) \!\cap\! \mathbf{E}(\mathbf{D_{xy}^{\star}},\sigma_{\mathrm{ref}};\mathbf{x}) \!\neq\! \{ \} \Big\}
\end{align}
of $\mathbf{S_{\nsucceq}}(\mathbf{D_y^{\star}}, \mathbf{y_{ref}})$. Here we use $\mathbf{H}(\mathbf{s})$ to denote the points contained in the $n$-dimensional hyperbox spanned by the points in $\mathbf{s}$ and $\mathbf{E}(\mathbf{D_{xy}^{\star}},\sigma_{\mathrm{ref}};\mathbf{x})$ to denote the points contained in the hyperellisoid with centers $\mu_1, \dots, \mu_n$ and eccentricities $\sigma_{\mathrm{ref}} \sigma_1, \dots, \sigma_{\mathrm{ref}} \sigma_n$, respectively.\par
This approximation effectively allows us to truncate the sum in \cref{eqn:app:evi} and therefore skip the calculation of the corresponding integrals which have a neglectable effect on the outcome.

\subsection{Probability of being non-dominated} \label{sec:app:nondomprob}
In this section, we outline the calculation of \cref{eqn:utility:SR:pnd}. Similar calculations can also be found in \cite{emmerich2005}. For a regression model with separable probabilites, \cref{eqn:py,eqn:app:py:separable}, we can write
\begin{align} \label{eqn:app:pnd:def}
p_{\nsucceq}(\mathbf{D_{xy}^{\star}};\mathbf{x}) & = p(\forall\, \mathbf{y} \in \mathbf{D_y^{\star}} \,\exists\, i \,:\, \hat{y}_i(\mathbf{D_{xy}^{\star}};\mathbf{x}) < y_i) \nonumber \\
& = \prod_{\mathbf{y} \in \mathbf{D_y^{\star}}} \sum_I \! \Big[ \prod_{l \in I} p(\hat{y}_i(\mathbf{D_{xy}^{\star}};\mathbf{x}) \!<\!  y_i) \nonumber \\
& \hspace{1.965cm} \prod_{l \notin I} p(\hat{y}_i(\mathbf{D_{xy}^{\star}};\mathbf{x}) \!\geq\! y_i) \Big]
\end{align}
with $I$ representing all possible combination of index sets of $\{1,\dots,n\}$ with at least one element. \Cref{eqn:app:pnd:def} can also be written as
\begin{align}
p_{\nsucceq}(\mathbf{D_{xy}^{\star}};\mathbf{x}) = \prod_{\mathbf{y} \in P(\mathbf{D_y^{\star}})} \sum_{b \in B} \prod_{l=1}^n p_i^{b_i}(\mathbf{D_{xy}^{\star}};\mathbf{y},\mathbf{x}).
\end{align}
Here we have introduced
\begin{align}
p_i^{b_i}(\mathbf{D_{xy}^{\star}};\mathbf{y},\mathbf{x}) \equiv \begin{cases} \int_{-\infty}^{y_i} \hat{p}_{y_i}(y_i | \mathbf{D_{xy}^{\star}},\mathbf{x}) & \text{if} \ b_i = 0 \\ \int_{y_i}^{+\infty} \hat{p}_{y_i}(y_i | \mathbf{D_{xy}^{\star}},\mathbf{x}) & \text{else} \end{cases}
\end{align}
and
\begin{align}
\mathbf{y_{\neq i}} \equiv (y_1, \dots, y_{i-1}, y_{i+1}, \dots, y_n),
\end{align}
and the set
\begin{align}
B \equiv \Big\{ (b_1, \dots, b_n) \,|\, & b_n \in \{0,1\} \,\forall\, i = 1 \dots n  \nonumber\\ 
	& \land\, \sum b_{i=1}^n > 0 \Big\}
\end{align}
containing all combinations of $\{0,1\}$ of length $n$ with repetition and with order-significance.\par
For a regression model with a normal probability density, \cref{eqn:pv:Gauss}, this expression can be straightforwardly rewritten in a closed form in terms of error functions. Specifically, one has
\begin{subequations}
\begin{align}
& p(\hat{y}_i(\mathbf{D_{xy}^{\star}};\mathbf{x}) \geq y_i) = \frac{1}{2} \left( 1 - \erf \left[ \frac{y_i - \mu_i}{\sqrt{2} \sigma_i} \right] \right)
\end{align}
and
\begin{align}
& p(\hat{y}_i(\mathbf{D_{xy}^{\star}};\mathbf{x}) < y_i) = \frac{1}{2} \left( 1 + \erf \left[ \frac{y_i - \mu_i}{\sqrt{2} \sigma_i} \right] \right),
\end{align}
\end{subequations}
respectively. In analogy to \cref{eqn:I:delta}, the limit \cref{eqn:pv:Gauss:sigma:vanish} can be used to obtain
\begin{subequations}
\begin{align}
p(\hat{y}_i(\mathbf{D_{xy}^{\star}};\mathbf{x}) \geq y_i) \xrightarrow{\sigma_i \goesto 0} \Theta \left[ \mu_i - y_i \right]
\end{align}
and
\begin{align}
p(\hat{y}_i(\mathbf{D_{xy}^{\star}};\mathbf{x}) < y_i) \xrightarrow{\sigma_i \goesto 0} \Theta \left[ y_i - \mu_i \right]
\end{align}
\end{subequations}
in case of vanishing standard deviations $\sigma_i$.

\end{document}